\documentclass[a4paper]{article}

\title{Parametric external predicates for the DLV System}
\author{G. Ianni, F. Calimeri, A. Pietramala, M.C. Santoro}

\def\ecom{{\tt \#}}
\newcommand\nop[1]{}

\begin{document}
\maketitle
\begin{abstract}
This document describes syntax, semantics and implementation
guidelines in order to enrich the DLV system with the possibility
to make external C function calls. This feature is realized by the
introduction of "parametric" external predicates, whose extension
is not specified through a logic program but implicitly computed
through external code.
\end{abstract}

\section{Intuition}

It is very usual the necessity to encode and employ external
functions in dlv. For instance, dlv lacks of mathematical
functions and string functions such as:

\verb+div(X,Y,Z)+, or \verb+sqr(X,Y)+.

...

\section{Syntax}
We introduce in dlv programs a new type of predicate which we will
call {\em external} predicate. The name of an external predicate
is preceded by the conventional \ecom\ symbol. For instance,

\ecom\verb+fatt(X,N)+, \ecom\verb+sum(1,2,3)+, are (external)
atoms with external predicate name. An external atom can appear
inside bodies and constraints only. An external predicate cannot
be true negated. It can be negated with default negation, and, in
this case, its variables must be safe in ordinary way.

\section{Semantics}

Roughly speaking, the formal semantics of an external atom is the
following. Given an external predicate \ecom\verb+p+ of arity $n$,
it is associated to it an "oracle" boolean function $p'$ taking
$n$ constant arguments. A ground atom \ecom\verb+p(a1,...,an)+ is
true iff $p'(a1,...,an)$ is true.

Example: consider the external predicate \ecom\verb+fatt+
implementing the external computation of the factorial function.
Its oracle function $fatt'$ is such that $fatt'(a,b)$ is true if
$b$ is the factorial of $a$.

\section{Operational semantics}
From the operational point of view, external predicates must be
implemented at grounding level. It is given a rule

\[
\verb+H(...) :- A1(...), ... , #P(X1,...,XN), ..., AN(...).+
\]

containing the external atom \ecom\verb+P+ associated to its
oracle $P'$. If \verb+X1,...,Xn+ are safe, external predicates are
easily implemented at grounding level by checking if
$P'(a1,...,an)$ is true for a given set of constants. In this
case, it suffices to reorder rules and put external atoms at the
very right edge of any rules where they appear. It is very
important instead to introduce an operational semantics taking
into account the possibility that some of the \verb+X1,...,XN+ are
not known.

For instance, assume it is given the external predicate
\verb+#sqr(N,S)+, computing the square $S$ of a given number $N$,
we may want the following rule to be possible

\verb+squares(S) :- number(N), #sqr(N,S).+

where $S$ is not bounded. {\em Note that, as a important
collateral effect, $S$ is not bounded at all, neither by an
\verb+#int+ atom. This works around the current need to ground and
represent all the integer constants. New integer constants are
introduced "on demand". Furthermore, this should allow to deal
with floating point values and strings in an efficient way. Only
constants actually necessary will be computed.}

In the above case, the call to oracle function $sqr'$ is not
feasible, since we do not know a priori which values of $S$ are
needed in order to call it.

In order to provide this further feature, we introduce the concept
of "talkative oracle".

It is given an external predicate \ecom\verb+p+, of arity $n$ and
its oracle function $p'$. A {\em pattern} is a list of $a$'s and
$A's$. A $a$ will represent a placeholder for a constant, whereas
an $A$ will be a placeholder for a variable. Given a list of
terms, the corresponding pattern will be given by replacing each
constant with $a$, and each variable with $A$. For instance the
pattern for the list of terms \verb+X,b,Y+ is $A,a,A$. Let $pat$
be a pattern of length $n$ having $k$ placeholders $a$ (which we
will call input positions), and $n-k$ placeholders of $A$ type
(which we will call output positions). A talkative oracle
$p^{pat}$ for the pattern $pat$, associated to the external
predicate \ecom\verb+p+, is a function taking $k$ constant
arguments, returning a tuple of arity $n-k$. $p^{pat}$ is such
that $p^{pat}( a_1, ..., a_k ) = b_1, ..., b_{n-k}$ iff $p'(
X^{pat} )$ is true, where $X^{pat}$ is a list of terms built from
$pat$ by in each input position a $a_i$, and each output position
with a $b_i$.

Given an external predicate \ecom\verb+p+, it may be associated
with one or more "consistent" talkative oracles. For instance,
consider the \ecom\verb+fatt+ external predicate. We associate to
it two talkative oracles, $fatt^{a,A}$ and $fatt^{A,a}$. For
instance,

\begin{eqnarray}
fatt^{a,A} ( 3 ) & = & 6 \\
fatt^{A,a} ( 6 ) & = & 3
\end{eqnarray}

consistently with the fact that $fatt'(3,6)$ is true.

From now on, given an external predicate, we will assume it comes
equipped with its oracle and a set of consistent talkative
oracles.

\section{Body reordering and safeness constraints}
At this point we are able to relax the constraint on safeness of
external atoms. A variable $Xi$, belonging to an external atom
\verb+p(X1,...,Xn)+ may be kept free if $p$ has a talkative oracle
associated to a pattern where $Xi$ is in the same position of a
$A$ symbol. For instance, assume \ecom\verb+sqr+ comes with the
$sqr^{a,A}$ oracle, and \ecom\verb+fatt+ comes with the
$fatt^{a,A}$ oracle.

The following rules are safe:

\verb+H(S) :- number(N), #sqr(N,S).+

\verb+H(S1) :- number(N), #fatt(N,S), #sqr(S,S1).+

The following are, viceversa, unsafe:

\verb+H(S) :- number(S), #sqr(N,S).+

\verb+H(S1) :- number(S), #fatt(N,S), #sqr(S,S1).+

\paragraph{Safeness rules}

A completely defined rule (or constraints), is a rule where the
order of instantiation has been completely fixed, from left to
right, and for each external atom \verb+#p+ an unique talkative
oracle has been chosen.

We consider three notion of safety:

\begin{itemize}
\item
Usual safeness. Each variable must appear in at least a positive
atom within the body.

\item
Weak safeness. This notion applies only to a completely defined
rule. A variable is weakly safe if it is either,
\begin{itemize}
\item
usually safe, or

\item
it appears in an external atom in output position (with respect to
the corresponding talkative oracle), and all the variables in
input position are weakly safe.
\end{itemize}

\item
Strong safeness. The strong safeness applies only in those
specific cases where recursion may induce an infinite Herbrand
universe. A rule will require strong safety if in the rules'
dependency graph (the one where rules appears as nodes) it appears
in some cycle. A variable $X$ is strongly safe if it is weakly
safe. In case $X$ appears within the head of the rule, then $X$
must be usually safe.
\end{itemize}

From now on we will use the notion of weak safety as default
safety constraint. Weak safety coincides with usual safety in case
external predicates are not defined. We require strong safety
whenever needed.

\paragraph{Strong safeness constraints}

There are cases where the above "weak" safety rules may lead to
the generation of an infinite Herbrand universe when employed in
recursive rules. Imagine to define an oracle for the \verb+#succ+
such that \verb+#succ(a,b)+ is true if and only if $a$ and $b$ are
integers and $b = a+1$. The grounding algorithm launched on the
rule

\verb+int(X) :- int(Y), #succ(X,Y).+

never reaches a fixed point. To avoid such cases, we introduce the
"strong" safety criterium which has to be applied only when
recursion is detected.

\paragraph{Body reordering}

In general, given a rule, the following tasks have to be performed:

\begin{enumerate}
\item
choice a suitable talkative oracle for each external atom in it,
such that the rule is weakly safe, and , if necessary, strongly
safe; intuitively, this problem is, in its general setting,
NP-complete (reduction from covering), but in practice there will
be very few talkative oracles for each external atom, so the
search space will be actually very small. Please note that there
could be more than one solution. We do not discuss at the moment
the problem of making the best choice. Just pick a suitable
talkative oracle for each external atom.

\item
put each external atom in a proper position such that:
\begin{itemize}
\item[a.] the rule rests safe;
\item[b.] the atoms ordering is efficient as much as possible.
\end{itemize}
\end{enumerate}

The two tasks are actually strictly related. Let's try to briefly
explain a possible algorithm to solve the problem.

Since the external built-in atoms are, in a sense, functionally
conceived (i.e. given a set of terms as ''input'' the built-in
will have return an unique value for the output positions) they
can be computed ''on-the-fly'', each time the grounding module
reaches them processing a body of some rule; and so it is
preferable to put them as soon as possible. This can also give a
way to choose the proper oracle per each atom: each time an
"ordinary" atom is placed, it is enough to check per each external
atom whether there are some oracles whose ''calling patterns'' are
subsets of the variables which are currently bounded. If so, the
external atom can be placed.

Imagine to being ordering the following rule: \\
\verb+     p(X) :- q(X, Y), s(Y, T), m(Z), n(Z, T), #r(Y, Z, T).+ \\
and let the ''r'' atom have the only oracle: \\
\verb+     #r: X,Y -> Z. + \\
Now let's be in the partial situation: \\
\verb+     p(X) :- m(Z), q(X, Y), ...???... + \\
The currently bounded variables are \{Z,X,Y\}. There is an oracle
for \#r with calling signature requiring \{Y,Z\}: so the atom
\#r(Y, Z, T) can be put just after those currently placed. \\

Please note as the previous example is trivial. In general, two
kinds of problem may arise:
\begin{itemize}
\item[a.] more than one oracle are suitable for an extern atom;
\item[b.] more than one atom are placeable at a given step.
\end{itemize}
So some criteria in order to choose between oracles has to be
depicted; and well as some criteria for choosing the order between
external atoms.

\paragraph{Grounding stage modification tips}

I suggest to introduce a derived class of atoms having the
possibility to call the selected talkative oracle whenever it is
necessary to discover new values. This atom may follow a caching
schema in order not to call the oracle when some call has already
been computed.

A supporting class for this new type of atoms could be:

\begin{verbatim}
class externalpredicate {

    CollectionOfOracles o;
    CollectionOfGroundAtoms g; // Caches already computed ground
                               // atoms. Corresponds to the
                               // current predicate extension.
    CollectionOfOracleCalls c; // Stores already called patterns.
    friend operator< (call v1, call v2);
    CollectionOfGroundAtoms callanoracle( oracle o, call v )
    {
        if (there is no call v1 <= v in c)
            push o.invoke(v) in g;
    }
}
\end{verbatim}

a call is simply a list of terms. There is a partial order between
calls, given by their generality. For instance the call
\verb+(a,b,X)+ precedes \verb+(a,b,c)+, whereas \verb+(c,d,X)+ and
\verb+(a,d,X)+ cannot be compared. Each talkative oracle is able
to respond to a given pattern, so each call must follow such a
pattern. An external atom in a rule has ONLY one talkative oracle
which can be called. It is anyway possible that the same external
predicate appears in the same rule or in another point of the
program and ANOTHER talkative oracle has been associated to it.
The class \verb+externalpredicate+ concentrates the oracle calls
in a central data structure, and performs oracle calls only if
necessary. An oracle call must be performed in the InstantiateRule
stage on demand.

Here: modifications on the instantiaterule.

\section{External predicate oracle specification language}

First of all, we need to provide a public header file
(\verb+extpred.h+) containing some public classes, coupled with a
binary dynamic library \verb+extpred.so+. The external predicates
programmer should include this header in order to program his/her
own external predicates.

\verb+extpred.h+ should contain:
\begin{itemize}
\item The definition of a public \verb+CONSTANT+ class. It should
be a minimal version of the \verb+TERM+ internal class. The
\verb+TERM+ class should be extended with constructors and methods
in order to convert to and from the \verb+CONSTANT+ class. It is
not necessary that \verb+CONSTANT+ is aware of \verb+TERM+ at all.

{\bf FIXME Another possibility is to do not rely on the TERM
class, but only have a {\em data} field and proper conversion
methods in order to convert from/to internal dlv types. The latter
are essentially only numbers (integers) and strings (char$*$). We
will talk about that a little bit later.}
\end{itemize}

The custom code should contain a set of functions taking as
argument an \verb+int+ representing the arity of the built-in
predicate and an array of CONSTANTs (each one of these is linked
to a term of the predicate).

Each function name will encode the name of the associated external
predicate and the pattern to which it responds to. Some of the
CONSTANTs on the array constitute the ''input'' and some others
the ''output'', accordingly to the specific pattern. A \verb+bool+
is returned specifying whether the ''unification'' has succeeded
or not.

To do that we suggest to introduce the preprocessor directive \\

\verb+#define BUILTIN(name,pattern)+ \\
\verb+           bool __name__pattern(CONSTANT *argv[])+ \\

where \verb+name+ is the external predicate name, and
\verb+pattern+ is the pattern of the oracle to be implemented.
Pattern is essentially a string constituted by ''i'' (lowercase)
and ''O'' (uppercase); the succession is positional, and the
meaning is that at j-th position of the pattern string will appear
an ''i'' if for the oracle being defined the j-th term is in
''input''; and as it can be easily imagined an ''O'' means that
the j-th term is in output. Let's here recall that this will mean
that each term for which an ''i'' is specified has to be bound
when the oracle is invoked (and we think to proper reorder the
body of the rule in order to ensure that).

Example:

\begin{verbatim}
// sketch of a external predicate definition source code
# include "extpred.h"

BUILTIN(fatt,iO)  // oracle for the pattern (a,A)
{
// We know we want just ONE parameter as input
assert (argc == 2);

// Convert the CONSTANT data to what we do have to compute
int x = argv[1].data;

// Just perform the computation...
assert ( x >= 0);
int fatt = 1;
if ( x > 0)
    for (int i = 1; i <= x; i++)
        fatt *= i;

// Convert the result into the proper CONSTANT data
CONSTANT result (fatt);

// Save the result onto the output vector
output.push_back(result);
}


BUILTIN(fatt,Xx)  // oracle for the pattern (A,a)
{
// We know we want just ONE parameter as input
assert (input.size == 1);

// Convert the CONSTANT data to what we do have to compute
int x = int( (input.begin().data()) );

// Just perform the computation...
assert ( x >= 0);

int base = 1, y = x;
int carry = 0;

bool ok = true;

while ( y > 1 && (carry == 0) )
    {
    y = x / base;
    carry = x % base;
    }

CONSTANT result;


// FIXME: DEFINE HOW TO REPRESENT FAILING (no unification)
if ( carry == 0 )
    result = base
else
    result = "FAIL";

output.push_back(result)

// Save the result onto the output vector
output.push_back(result);
}
\end{verbatim}

Some notes:

\begin{itemize}
\item[.]
we do have to finely declare how to treat input and output (i.e.:
should the size have to equal the number of ''input arguments'' of
the oracle? If not, the oracle may take the first arguments, or
the latest ones...
\item[.]
the same for output vector: can we simply care of inserting it
from back?
\item[.]
how to manage the data contained in each CONSTANT? We should have
to precisely project the class, since the types have to be {\em at
least} real constants or numbers...
\item[.]
how to model class CONSTANT in order to represent even the failing
case (i.e., the corresponding missing unification for a
traditional predicate);
\item[.]
what else?
\end{itemize}

\section{Using Externally Defined Built-ins into dlv Programs}
Once the user has her own libraries and wants to use one or more
built-ins into a logic program, she has simply to tell dlv where
to find oracle definitions (i.e., where are the (compiled)
libraries. The syntax is quite the same as for object oriented
languages, and requires the use of the \verb+#include+ directive.
The directive can ask to import a complete ''package'' (i.e.,
everything is in a folder) or a single library file:\\

\verb+ #include mylib.strings.*+\\

imports every lib file contained into strings package, while\\

\verb+ #include mylib.strings.compare+     $<$-- FIXME: which extension?\\

imports only the compare library.\\
From then, the program can exploit all built-ins defined into the
imported libraries. For instance, let the \verb+#contains+
built-in be defined in some imported library:\\

\verb+ p(X) :- q(_,X), #contains(X,"stripes").+ \\

NOTE: the default path should be \verb+./lib+. A command line
option should be provided in order to specify a different path
(folders separated by semicolons): \\

\verb+ .... -path=/usr/local/lib/dlv;/home/myuser/lib;+\\

\subsection{Some Cares}
\begin{itemize}
\item[.]
In case of predicate name conflict, the last import directive
takes precedence, but it is suggested to output a warning in this
case. The wanted version of a given predicate may be explicitly
selected using an external atom in the form
\verb+#packagename.predicatename(Vars)+ inside the logic program.
\item[.]
Please note as more than one \verb+#import+ directive can be
given, but all of them have to be put at the {\bf beginning of the
file} containing the logic program.
\item[.]
The user can freely define ''standard'' predicates having the same
name of some built-in. For instance the same program may contain
\verb+p(...)+ and \verb+#p(...)+, and they will be distinguished.
\item[.]
On the other hand, the user cannot override some identifiers;
think about predefined aggregates (\verb+#sum, #count,+ etc.) or
keywords like \verb+#template+.
\end{itemize}

\section{Tips on dynamic linking of external built-ins}
We suggest two ways of linking new built-ins: either statically
(need recompiling DLV), or dynamically (recompiling is not
needed). The second way allows third parties to develop and
introduce new built-in libraries for dlv, and it is recommended.

\section{Notes on wrapping of the older builtins ( arithmetics and so on )}

\section{Some Notes on Implementation}

\begin{enumerate}
\item
During rules' reordering (see {\em orderedBody} in
{\em grounding-body.h}) the function {\em isAdmissible\footnote{Or
''lookup'', we'll see.}} is invoked. It is a boolean function
receiving a built-in predicate ad the list of bounded variables;
it checks whether there exists a suitable oracle for the current
built-in, and if so it returns true. Please note that (in case of
success) it also sets a pointer (or an index, let's see) to the
found oracle, after a safety check.

\item
A brand new class called \verb+EXTENDED\_BUILTIN+ will be
built. It will have to be able to manage built-ins replacing the
existing ones (''standard'') as well as the new (''extended''). A
possible (rough) structure may be the following:

\begin{verbatim}
    class EXTENDED_BUILTIN
        {
        vector<ORACLE> *oracle;

        ...........................

        bool isAdmissible(const char* name, const TERMS *bounded)
        };
\end{verbatim}

The \verb+oracle+ vector stores all existing oracles related to
the single built-in (i.e. all user-defined oracles in the dynamic
library).

\item
Essentially an instance of this class has to built per each
built-in, once and for all built-in atoms related to it. All
these, will refer to this.

All instances of \verb+EXTENDED\_BUILTIN+ will be stored in an
(hash?)map having the name as key.

\item
Each instance of a built-in atom should have a pointer to the
actual oracle.

\item
The \verb+CONSTANT+ class have to provide suitable methods in
order to cast to standard types, such as
\verb+toString()+, \verb+toInt()+, etc. \\
Ex.: \verb+ int base = argv[2].toInt()+

\item
In order to call the proper oracle we can use a \verb+switch+ for
rapid prototyping, and then use a more efficient array of pointers
to functions.

\item
During the grounding phase it is possible that some oracle
generates {\em new} constants (i.e., not yet present in the
Herbrand Universe): think about some ''output'' constant. If so,
we do have to maintain up-to-date the \verb+CONSTANT\_NAMES+
collection (i.e., the ''register'' of existing constants). Maybe
we will need proper methods.

NOTE: In such a case, a problem about generation of an infinite
Universe arises. We could:

\begin{itemize}
\item
Check under a safety criterion (see above).
\item
Make the user responsible about that.
\end{itemize}

\item
Does the \verb+#maxint+ setting have to be forced in case of
extended built-ins usage?

\item
In a successfully grounded rule a built-in atom will be always
true, so it can be deleted. Check whether it may be automatically
removed or a new procedure is needed.

\item
It is mandatory to define per each built-in the ''base'' oracle
(i.e., the one with all parameters bounded -- iii [...] i).
\end{enumerate}

\section{Ambiguity on choosing the oracle}
During the body reordering the built-ins are placed and a proper
oracle has to be chosen among all admissible (see above). In case
more than one oracle are admissible, a way to choose has to be
found. Since built-ins are conceived under ''functional''
semantics we could take any of the oracles: the result will be
always the same. For the sake of efficiency, we think that the
less are the output \verb+CONSTANTS+, the more is preferable the
oracle: in fact, this will allow to avoid more checks with already
bounded variables.

NOTE: should it be necessary to ask the user for a partial order
between oracles?

\end{document}